\documentclass{article}
\usepackage{ijcai19}

\usepackage{algorithm}
\usepackage{algpseudocode}
\usepackage{amsmath}
\usepackage{amssymb}
\usepackage{array}
\usepackage{balance}
\usepackage[pangram]{blindtext}
\usepackage{booktabs,threeparttable}
\usepackage[small]{caption}
\usepackage{color}
\usepackage{comment}
\usepackage{dblfloatfix}
\usepackage[inline]{enumitem}
\usepackage{epsfig}
\usepackage{etoolbox}
\usepackage{fancybox}
\usepackage{filecontents}
\usepackage{fixltx2e}
\usepackage{float}
\usepackage{graphicx}
\usepackage[hidelinks]{hyperref}
\usepackage[utf8]{inputenc}
\usepackage{lastpage}
\usepackage{lipsum}
\usepackage{listings}
\usepackage{multirow}
\usepackage{multicol}
\usepackage{pifont}
\usepackage{placeins}
\usepackage{rotating}
\usepackage{setspace}
\usepackage{soul}
\usepackage[hang, tight]{subfigure}
\usepackage{threeparttable}
\usepackage{times}
\usepackage{url}
\usepackage{verbatim}
\usepackage{wrapfig}
\usepackage[usenames,dvipsnames]{xcolor}
\usepackage{color}
\newbool{inccomment}
\setbool{inccomment}{true}
\newcommand{\XX}[1]{\ifbool{inccomment}{{\color{magenta} #1}}{}}
\newcommand{\CT}[1]{\ifbool{inccomment}{{\color{magenta}CT\@: #1}}{}}
\newcommand{\NT}[1]{\ifbool{inccomment}{{\color{blue}NT\@: #1}}{}}
\newcommand{\TD}[1]{\ifbool{inccomment}{{\color{orange}#1}}{}}
\newcommand{\FN}[1]{\ifbool{inccomment}{{\color{OliveGreen}#1}}{}}
\newcommand{\GR}[1]{\ifbool{inccomment}{{\color{Tan}#1}}{}}
\newcommand{\LD}{\ifbool{inccomment}{{\color{magenta}\\============================================\\}}}
\newcommand{\RF}{\ifbool{inccomment}{{\color{green}~[R]}}}

\title{Interpreting and Evaluating Neural Network Robustness}

\author{
	Fuxun Yu$^1$, Zhuwei Qin$^2$, Chenchen Liu$^3$, Liang Zhao$^4$, Yanzhi Wang$^5$, Xiang Chen$^6$ \\
	$^{1,2,4,6}$George Mason University,
	$^{3}$Clarkson University,
	$^{5}$Northeastern University\\
	chliu@clarkson.edu$^3$,  yanz.wang@northeastern.edu$^5$ \\
	\{fyu2, zqin, lzhao9, xchen26\}@gmu.edu$^{1,2,4,6}$ \\
}
\normalsize

\begin{document}
\maketitle
\vspace{-8mm}
\begin{abstract}
Recently, adversarial deception becomes one of the most considerable threats to deep neural networks.
	However, compared to extensive research in new designs of various adversarial attacks and defenses, the neural networks' intrinsic robustness property is still lack of thorough investigation.
	This work aims to qualitatively interpret the adversarial attack and defense mechanism through loss visualization, and establish a quantitative metric to evaluate the neural network model's intrinsic robustness.
    The proposed robustness metric identifies the upper bound of a model's prediction divergence in the given domain and thus indicates whether the model can maintain a stable prediction.
    With extensive experiments, our metric demonstrates several advantages over conventional adversarial testing accuracy based robustness estimation:
	(1) it provides an uniformed evaluation to models with different structures and parameter scales;
	(2) it over-performs conventional accuracy based robustness estimation and provides a more reliable evaluation that is invariant to different test settings;
	(3) it can be fast generated without considerable testing cost.
\end{abstract}

\section{Introduction}
\label{sec:intro}
In the past few years, Neural Networks (NNs) have achieved superiors success in various domains, \textit{e.g.}, computer vision~\cite{computervision}, speech recognition~\cite{speechrecognition}, autonomous systems~\cite{autodriving}, \textit{etc}.
	However, the recent appearance of adversarial attacks~\cite{fgsm} greatly challenges the security of neural network applications:
		by crafting and injecting human-imperceptible noises into test inputs, neural networks' prediction results can be arbitrarily manipulated~\cite{advtrain}.
	Until now, the emerging pace, effectiveness, and efficiency of new attacks always take an early lead to the defense solutions~\cite{cw}, and the key factors of the adversarial vulnerabilities are still unclear, leaving the neural network robustness study in a vicious cycle.

In this work, we aim to qualitatively interpret neural network models' adversarial vulnerability and robustness, and establish a quantitative metric for the model-intrinsic robustness evaluation.
	To interpret the robustness, we adopt the loss visualization technique~\cite{goodfellow}, which was widely used in model convergence studies.
	As adversarial attacks leverage perturbations in inputs, we switch the loss visualization from its original parameter space into the input space and illustrate how a neural network is deceived by adversarial perturbations.
	Based on the interpretation, we design a robustness evaluation metric to measure a neural network's maximum prediction divergence within a constrained perturbation range.
	We further optimize the metric evaluation process to keep its consistency under extrinsic factor variance, \textit{e.g.}, model reparameterization~\cite{sharpminima}.

Specifically, we have the following contributions:
\vspace{-0.5mm}
\begin{itemize}
	\item We interpret the adversarial vulnerability and robustness by defining and visualizing a new loss surface called decision surface. Compared to the cross-entropy based loss surface, the decision surface contains the implicit decision boundary and provides better visualization effect;
		\vspace{-5mm}
	\item We testify that adversarial deception is caused by the neural network's neighborhood under-fitting. Our visualization shows that adversarial examples are naturally-existed points lying in the close neighborhood of the inputs. However, the neural network fails to classify them, which caused the adversarial example phenomenon;
		\vspace{-1mm}
	\item We propose a robustness evaluation metric. Combined with a new normalization method, the metric can invariantly reflect a neural network's intrinsic robustness property regardless of attacks and defenses;
		\vspace{-1mm}
	\item We reveal that under certain cases, \textit{e.g.}, defensive distillation, the commonly-used PGD adversarial testing accuracy can give unreliable robustness estimation, while our metric could reflect the model robustness correctly.
\end{itemize}
Extensive evaluation results show that our defined robustness metric could well indicate the model-intrinsic robustness across different datasets, various architectures, multiple adversarial attacks, and different defense methods.

\section{Background and Related Work}\label{sec:related}

\subsection{Adversarial Attacks and Robustness}
Adversarial examples were firstly introduced by~\cite{lbfgs}, which revealed neural networks' vulnerability to adversarial noises and demonstrated the gap between the artificial cognition and human visual perception.
	Since then, various adversarial attacks were proposed, such as L-BFGS attack~\cite{fgsm}, FGSM attack~\cite{advtrain}, C\&W attack~\cite{cw}, black-box attack~\cite{blackbox}, \textit{etc}.

Driven by the appearance of adversarial attacks, corresponding defense techniques also emerged, including adversarial training~\cite{advtrain}, defensive distillation~\cite{distillation}, gradient regularization~\cite{aaai}, adversarial logit pairing~\cite{alp}, \textit{etc}.
	Among those, MinMax robustness optimization~\cite{minmax} is considered as one of the most potent defenses, which boosts model accuracy by integrating the worst-case adversarial examples into the model training.

Currently, testing accuracy under adversarial attacks is used to evaluate the model robustness.
	However, it is highly affected by the attack specifications and can't comprehensively reflect the actual robustness regarding model-intrinsic properties.
    For example, one commonly used way to evaluate the model robustness is adopting the testing accuracy under projected gradient descent (PGD) attack as an estimation.
	However, our experiments demonstrate that such a robustness estimation is highly unreliable: a model with a high PGD testing accuracy could be easily broken by other attacks.

In this work, we aim to provide an intrinsic robustness property evaluation metric that is invariant from the specifications of models, attacks, and defenses.

\subsection{Neural Network Loss Visualization}

Neural network loss visualization is considered as one of the most useful approaches in neural network analysis own to its intuitive interpretation.
	Proposed by~\cite{goodfellow}, the loss visualization is utilized to analyze model training and convergence.
	Later,~\cite{largebatch} further revealed that flat local minima is the key to model generalization in parameter space.
	However, a model reparameterization issue was identified by~\cite{sharpminima} that the model parameter scaling may distort the geometry properties.

In this work, we adopt the concept of the loss visualization to analyze the neural network's loss behaviors under adversarial perturbations.
	Meanwhile, we will also provide a normalization method to solve the model reparameterization problem and derive our scaling-invariant robustness metric.

\subsection{Visualization Space Selection}
Besides of solving the reparameterization issue, the loss visualization needs further customization for the adversarial perturbation analysis.
	As the loss visualization mainly evaluates a neural network's generalization ability, it focuses on the \textit{parameter space} to analyze the model training and convergence in previous works.
	However, such an analysis focus doesn't fit well in the adversarial attacks and defenses, whose action scope is in \textit{input space}.
	On the other hand, loss function in the \textit{input space} measures the network’s loss variations \textit{w.r.t} the input perturbations.
	It naturally shows the influence of adversarial perturbations and is suitable for studying the robustness to adversarial perturbations.
	Therefore, we extend the previous methods into the \textit{input space}.

\begin{figure}[!tb]
	\centering
	\hspace*{-3mm}\includegraphics[width=3.5in]{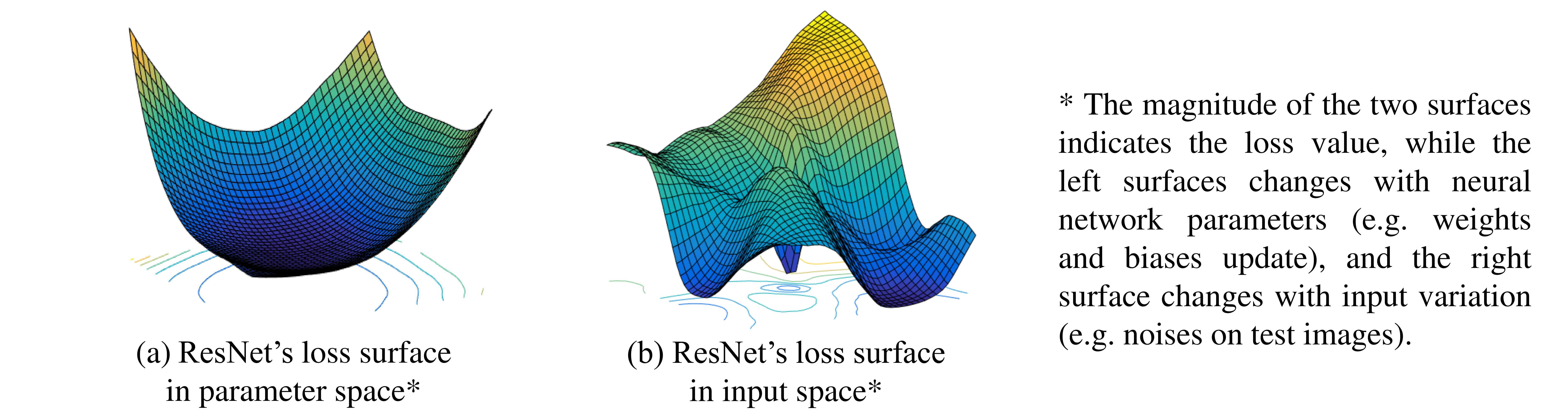}
	\vspace{-5mm}
	\caption{(a) ResNet's Loss Surface in Parameter Space (b) ResNet's Loss Surface in the Input Space: The loss surface demonstrates significant non-smooth variation.}
	\label{fig:1}
	\vspace{-3mm}
\end{figure}

Fig.~\ref{fig:1} showed two examples of the visualized loss surface of an ResNet model in the \textit{parameter space} and the \textit{input space}, which illustrate the difference between the two visualization spaces.
	Although the loss surface in the \textit{parameter space} can show a flat minima, its significant non-smooth variations in the \textit{input space} demonstrate the loss is highly sensitive to input perturbations, which can be adversarial vulnerabilities.
	In this work, we will adopt the \textit{input space} as the default visualization setting for robustness interpretation.

\section{Adversarial Robustness Interpretation}
\label{sec:interpret}

In this section, we will use the customized loss visualization to reveal the mechanism of adversarial perturbation and neural network robustness.

\subsection{Neural Network Loss Visualization }
\noindent\textbf{Loss Visualization Basis: }
	The prediction of a neural network can be evaluated by its loss function $F(\theta, x)$, where $\theta$ is the model parameter set (weight and bias) and $x$ is the input.
	As the inputs \textit{x} are usually constructed in a high-dimensional space, direct visualization analysis on the loss surface is impossible.
	To solve this issue, the loss visualization projects the high-dimensional loss surface into a low-dimensional space to visualize it (\textit{e.g.} a 2D hyper-plane).
	During the projection, two vectors $\alpha$ and $\beta$ are selected and normalized as the base vectors for $x$-$y$ hyper-plane.
	Given an starting input point $o$, the points around it can be interpolated, and the corresponding loss values can be calculated as:
\begin{equation}
	V(i, j, \alpha, \beta) = F(o + i \cdot \alpha + j \cdot \beta),
	\label{eq:1}
\end{equation}
where, the original point $o$ in the function $F$ denotes the original image, $\alpha$ and $\beta$ can be treated as the unit perturbation added into the image, and the coordinate $(i, j)$ denotes the perturbation intensity.
	In the loss visualization, a point's coordinate also denotes its divergence degree from the original point along $\alpha$ and $\beta$ direction.
	After sampling sufficient points' loss values, the function $F$ with high-dimensional inputs could be projected to the chosen hyper-plane.

\vspace{1mm}
\noindent\textbf{Decision Surface Construction: }
As the loss visualization is mostly used to analyze model convergence, the loss function $F(\theta, x)$ is usually represented by the cross-entropy loss, which constructs a conventional \textit{loss surface} in the visualization.
	However, one critical limitation of the cross-entropy based loss surface is that, it cannot qualitatively show the explicit decision boundary of one input test, and less helpful for adversarial deception analysis.

Therefore, we propose a \textit{decision surface} to replace the \textit{loss surface} in the loss visualization:
\begin{equation}
	S(x) = Z(x)_t - max\{ Z(x)_i,~i \neq t\},
	\label{eq:2}
\end{equation}
where, $Z(x)$ is the logit output before the softmax layer, and $t$ is the true class index of the input $x$.
	The decision function $S(x)$ evaluates the confidence of prediction.
	In the correct prediction cases, $S(x)$ should always be positive, while $S(x)<0$ denotes a wrong prediction.
	Specifically, $S(x) = 0$ indicates the equal confidence for both correct and wrong prediction, which is the \textit{decision boundary} of model.
	Consequently, the visualization surface constructed by the function $S(x)$ is defined the decision.
	Different from the cross-entropy based \textit{loss surface}, the \textit{decision surface} demonstrates explicit decision boundaries, and assist the adversarial  analysis.

\subsection{Visualizing Adversarial Vulnerability}

\textbf{Experimental Analysis: }
Based on the loss visualization, we project a neural network's loss behavior into 2D hyper-planes.
	By comparing the model's 4 different types loss behavior in \textit{decision surface}, we provide a experimental analysis for the adversarial vulnerability.

\begin{figure}[!tb]
	\centering
	\includegraphics[width=3.3in]{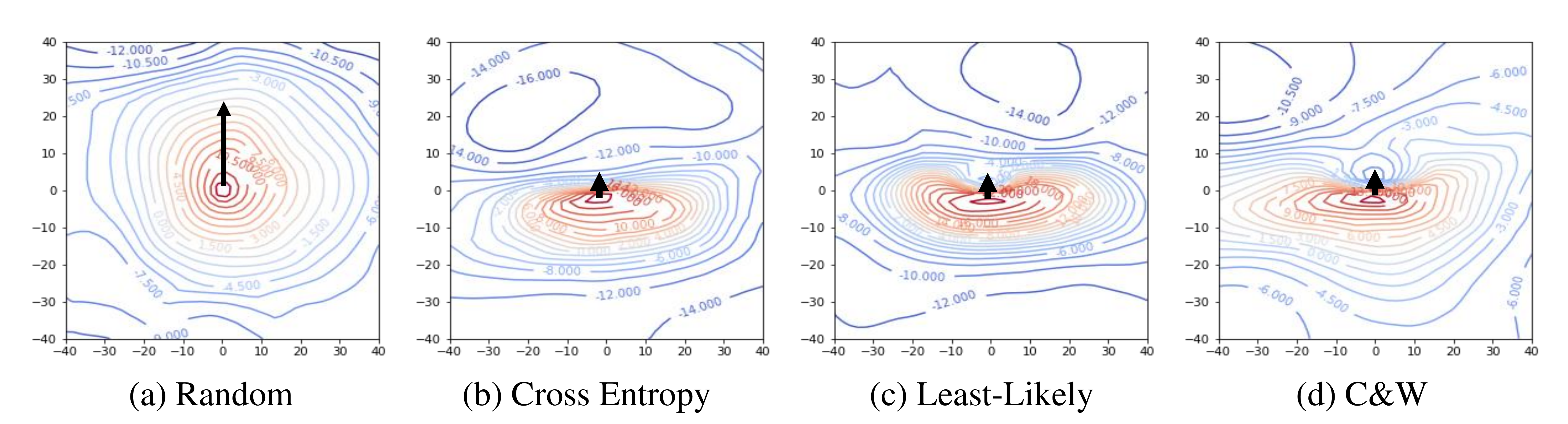}
		\vspace{-2mm}
	\caption{Adversarial vulnerability demonstration when loss surface in input space is projected onto different hyperplanes.}
	\label{fig:xent-cw}
	\vspace{-4mm}
\end{figure}

As shown in Fig.~\ref{fig:xent-cw}, the visualized hyper-planes have the central points ast the original neural network inputs, and their \textit{x}-axes share the same input divergence direction -- $\alpha$.
	Meanwhile, each hyper-plane has a dedicated input divergence direction -- $\beta$ along the \textit{y}-axis, which indicate 4 kinds of perturbations, including random noise, cross-entropy based non-targeted FGSM attack~\cite{fgsm}, Least-likely targeted FGSM attack~\cite{fgsm}, and non-targeted C\&W attack~\cite{cw}.
	Specifically $\beta$ values in the three adversarial attacks can be determined as:
\begin{equation}
	\begin{split}
		&\beta_0 = sign(\textit{N}(\mu=0, \sigma=1)), \\
		&\beta_1 = sign(- \nabla_x~y_t \cdot log(softmax(Z))), \\
		&\beta_2 = sign(\nabla_x~y_l \cdot log(softmax(Z))), \\
		&\beta_3 = sign(\nabla_x~max\{ Z(x)_i, i \neq t\}-Z(x)_t),
	\end{split}
	\label{eq:4}
\end{equation}
where $N$ is normal distribution, $Z$ is the logit output, $y_t$ is the true class label, $y_l$ is least likely class label (both one-hot).

%

In Fig.~\ref{fig:xent-cw}, we use arrows to show the shortest distance to cross the decision boundary $L(x)$=0.
	As projected in Fig.~\ref{fig:xent-cw}(a), when the input is diverged by the perturbation along a random direction, it will take much longer distance to cross the decision boundary.
	This explains the common sense that natural images with small random noises won't degrade neural network accuracy significantly.
	By contrast, for the adversarial attacks projected in Fig.~\ref{fig:xent-cw}(b)$\sim$(d), the attacks find aggressive directions ($\beta$ direction shown in \textit{y}-axis), towards which the decision boundary is in the close neighborhood around the original input. Therefore, adding those small perturbations that even human can't perceive into input can mislead the model decision and generates adversarial examples.

\vspace{1mm}
\noindent\textbf{Vulnerability Interpretation:}
The above experimental analysis reveals the nature of adversarial examples: Although a neural network seems to converge well after the model training (the demonstrated model achieves 90\% accuracy on CIFAR10), there still exist large regions of image points that the neural network fails to classify correctly (as shown by the large regions beyond the decision boundary in Fig.~\ref{fig:xent-cw}(b)$\sim$(d)).
	What's worse, some of these regions are extremely close to the original input point (even within $\ell_{\inf} < 1$ distance).
	
Base on these analysis, we could conclude that, rather than being ``generated" by attackers, the adversarial examples are ``naturally existed" already that models fail to learn correctly.
	To fix such intrinsic vulnerability of neural networks, the essential and ultimate robustness enhancement should focus on solving the ``neighborhood under-fitting" issue.


\subsection{Interpreting Adversarial Robustness}
To verify our geometric robustness theory, we compare two pairs of robust and natural models trained on MNIST and CIFAR10 respectively.
	These models are released from the adversarial attacking challenges \footnote{https://github.com/MadryLab/mnist\_challenge}\footnote{https://github.com/MadryLab/cifar10\_challenge}, and built with the same structure but different robustness degrees (natural training and MinMax training~\cite{minmax}).

\begin{figure}[!tb]
  \centering
  \includegraphics[width=3.3in]{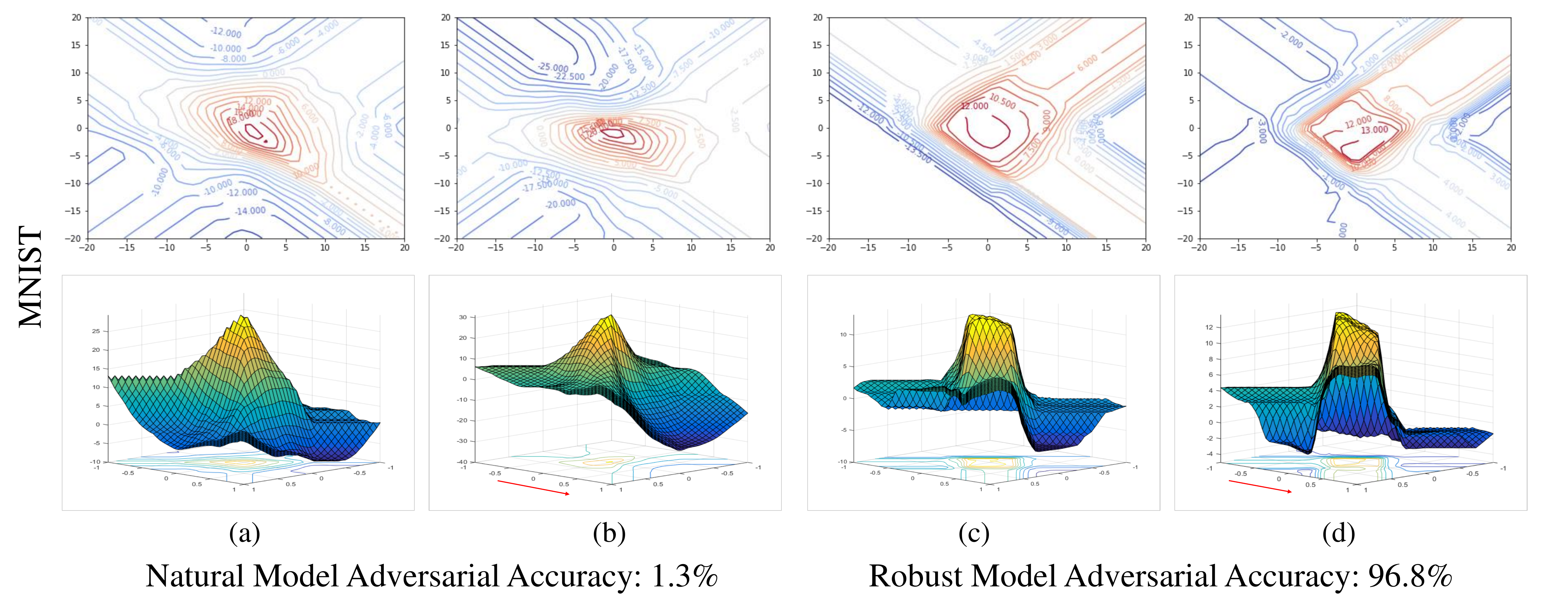}
  \caption{Decision surfaces of the natural and robust models on MNIST. (a)-(b): natural model surfaces in random and adversarial projection; (c)-(d): robust model surfaces in random and adversarial projection (each unit denotes 0.05 perturbation step size)}
  \label{fig:mnist}
\end{figure}

\begin{figure}[!tb]
  \centering
  \includegraphics[width=3.3in]{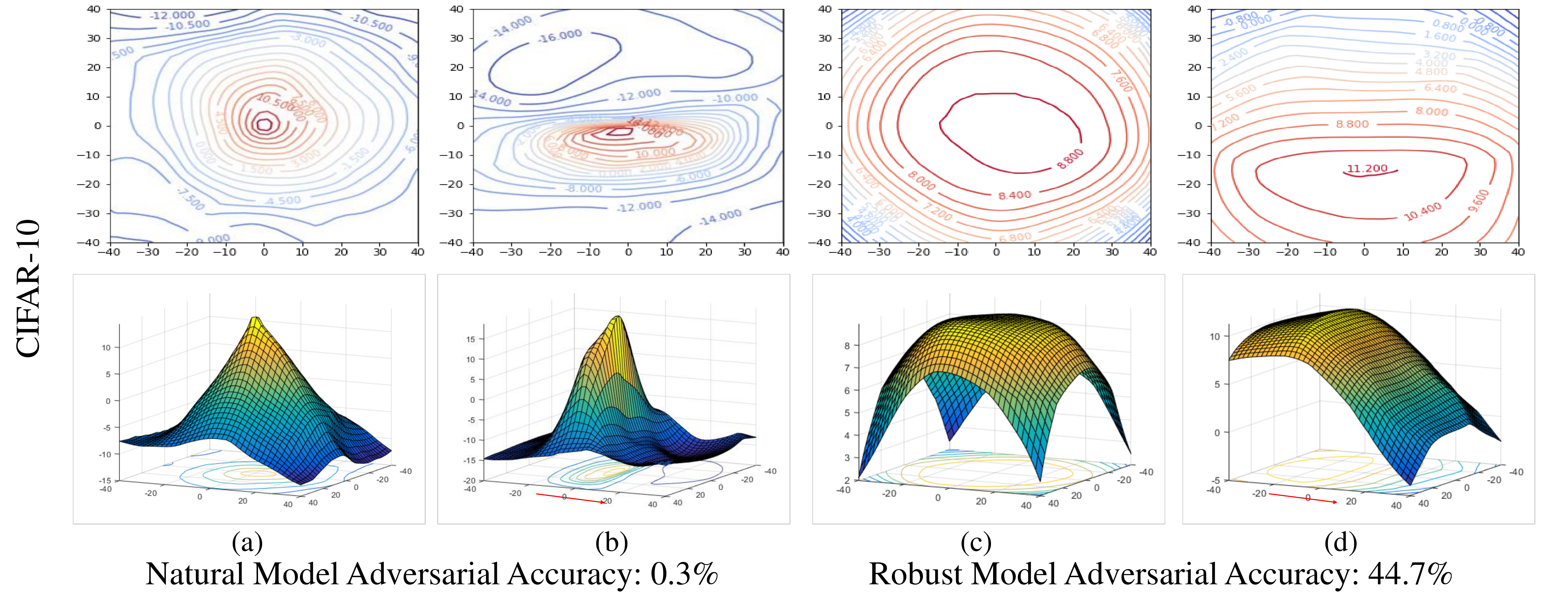}
  \caption{Decision Surface of natural and robust model on CIFAR10 (step size = 1). As assumed, natural model's surface shows sharp peaks and cliffs while robust model's shows flat plateau.}
  \label{fig:cifar}
 	\vspace{-4mm}
\end{figure}

	To verify our theory, we visualize the models' decision surfaces for interpretation:
	(1) As shown in Fig.~\ref{fig:mnist}, dramatic differences between the natural and robust decision surfaces can be observed: Natural (vulnerable) model's decision surfaces show sharp peaks and large slopes, where the decision confidence could quickly drop to negative areas (wrong classification regions).
	(2) By comparison, on robust decision surfaces (shown in Fig.~\ref{fig:mnist}(c)(d)), all neighborhood points around the original input point are located on a high plateau with $L(x) > 0$ (correct classification regions).
	(3) The surface in the neighborhood is rather flat with negligible slopes until it reaches approximately $\ell_\infty=0.3$ constraints, which is exactly the adversarial attack constraint used in robust training.
	Similar phenomenon could be observed in Fig.~\ref{fig:cifar} on CIFAR10.

Such robust model's loss geometry verifies our previous conclusion that, fixing the neighborhood under-fitting issue is the essential robustness enhancement solution for neural networks. And a flat and wide plateau around the original point on decision surface is one of the most desired properties of a robust model.

\section{Adversarial Robustness Evaluation}
\label{sec:evaluate}


\subsection{Formal Definition of Robustness Metric}
As aforementioned, the decision surface of a robust model should have a flat neighborhood around the input point $x$.
    Intuitively explanation is that a robust model should have good prediction stability--its prediction does not have significant change with small perturbations.
    In fact, models are not always robust--the predictions of a model on clean and noisy inputs are not always the same and can diverge to a large extent with small adversarial noises. 
    As such, \textit{given a feasible perturbation set, the maximum divergence between the original prediction and the worst-case adversarial prediction could be used to denote the model's vulnerability degree (i.e., the inverses of model robustness)}.

Based on this definition, firstly, we calculate the divergence between two predictions on an original input and an adversarial input with perturbations in a defined range. 
    Specifically, we use the Kullback–Leibler divergence, which is known as KL Divergence ($D_{KL}$) and is a common evaluation metric on measuring the divergence between two probability distributions. 
    The formal robustness could be estimated by:
\begin{equation}\label{eq:4}
\begin{split}
  \psi(x) = \frac {1} {\max\limits_{\delta \in set} D_{KL}(P(x), ~P(x+\delta))},
\end{split}
\end{equation}
where $P(\cdot)$ is the prediction results from the evaluated model. 
	A lower divergence $D_{KL}$ indicates the model is more robust as a more stable prediction is maintained.
    The final robustness metric $\psi(x)$ is defined inversely proportional to the maximum $D_{KL}$ since the largest divergence will generate the smallest robustness score $\psi(x)$.
To obtain the $\max$ term in Eq.~\ref{eq:4}, we use the gradient ascent algorithm to directly optimize the KL-divergence, which demonstrates accurate and stable estimations that we will show in Sec.~\ref{sec:exp}.

\subsection{Invariant Normalization against \\~\hspace{7mm}Model Reparameterization}

The robustness metric defined in previous works has a problem called ``model re-parameterization":  
    on the condition that weights and biases are enlarged by the same coefficients simultaneously, a neural network model's prediction results and its robustness property will not change, while the defined KL divergence can have dramatic change~\cite{sharpminima}. 

To solve this problem, we design a simple but effective normalization method: 
    the basic idea is to add a scale-invariant normalization layer after the logit layer output. 
    Since the neural network before the logit layer is piecewise-linear, we could then use normalization to safely remove the scaling effect of model reparameterization. 
The basic process is as follows: firstly, we attain a confidence vector of the logit layer, which can contain either positive or negative values; 
    then we divide them by the max-absolute-value to normalize the confidence vector to the range of (-1, 1) and re-center them into positive range (0, 2). 
    Owning to the max division, the final confidence vector will not change even when the parameters are linearly scaled up (or down).
    Finally, we use a simple sum-normalization to transform the confidence vector to a valid probability distribution. The overall normalization is:
\begin{equation}\label{eq:6}
  P(x) = \frac {\tilde{F}(x)}{\sum_{i}{\tilde{F}(x_i)}},  ~\tilde{F}(x) = \frac{F(x)}{\max |F(x)|}+1.
\end{equation}
Here $P(x)$ is the final normalized probability distribution, $\tilde{F}$ is the normalized confidence vector, $F(x)$ is the original logit layer output, and $x$ is the input. 
    By the above normalization method, we could successfully alleviate the model reparameterization effect, which is shown in Sec.~\ref{sec:exp}. 
\section{Robustness Evaluation Experiments}
\label{sec:exp}

In this section, we evaluate our proposed robustness metric in various experimental settings and compare it with conventional evaluation based on adversarial testing accuracy.

\subsection{Experiment Setup}
To test the generality of our metric for neural networks' robustness evaluation, we adopt three common datasets (\textit{i.e.} MNIST, CIFAR10, and ImageNet) and different models for the experiment, including FcNet, LeNet, ConvNet, ResNet18, ResNet152, and DenseNet.

  To further test our metric on neural networks with different robustness degrees, the following defense settings are applied: No Defense, Adversarial Training~\cite{fgsm}, Gradient Regularization Training~\cite{aaai}, Defensive Distillation~\cite{distillation}, Gradient Inhibition~\cite{wujie} and MinMax Training~\cite{minmax}\footnote{The gradient regularization and MinMax training is re-implemented with Pytorch, which may cause small deviations from the original reported accuracy.}.

  Correspondingly, the robustness verification is based on referencing the adversarial testing accuracies from two currently strongest attacks: 30-step PGD (PGD-30) attack based on cross-entropy loss and 30-step CW (CW-30) attacks based on C\&W loss.
  The adversarial perturbations are constrained by the $\ell_\infty$-norm as 0.3/1.0, 8.0/255.0, 16.0/255.0 on MNIST, CIFAR10, and ImageNet respectively.

\begin{table}[!tb]
  \centering
  \scriptsize
  \renewcommand\arraystretch{1.3}
  \setlength{\tabcolsep}{3.5mm}{
  \caption{Robustness Metric Evaluation on MNIST}
  \vspace{-2.5mm}
  \begin{tabular}{ccccc}
  \hline \hline
  Model & Defense  & $\psi$(x)  & \begin{tabular}[c]{@{}c@{}}PGD-30\\ Accuracy\end{tabular} & \begin{tabular}[c]{@{}c@{}}C\&W-30\\ Accuracy\end{tabular} \\ \hline \hline
  \multirow{3}{*}{FcNet} & No Defense     & \textbf{73.36} & 0.73\%                                                    & 0.2\%                                                    \\ \cline{2-5}
            & AdvTrain & \textbf{80.43} & 4.43\%                                                    & 2.12\%                                                   \\ \cline{2-5}
            & MinMax   & \textbf{297.2} & 82.9\%                                                    & 80.3\%                                                   \\ \hline
            \multirow{3}{*}{LeNet} & No Defense       & \textbf{93.8} & 2.82\%                                                    & 1.01\%                                                   \\ \cline{2-5}
            & AdvTrain & \textbf{264.7} & 51.8\%                                                    & 46.2\%                                                   \\ \cline{2-5}
            & MinMax   & \textbf{958.4} & 92.3\%                                                    & 90.3\%                                                   \\ \hline
\hline
\end{tabular}
*AdvTrain: \cite{fgsm}, MinMax: \cite{minmax}.
\label{table:mnist}}
  \vspace{-4.5mm}
\end{table}

\subsection{Robustness Metric Evaluation}
\textbf{MNIST Experiments: }
On MNIST dataset, the results are shown in Table~\ref{table:mnist}:
  (1) The results firstly demonstrate that our metric could well reflect different robustness degrees on the same neural network model.
  For example, three FcNet models show increasing robustness in $\psi(x)$, which aligns well with their reference accuracies from both PGD-30 and CW-30 attack;
  (2) The results also show the generality of our metric on FcNet and LeNet models.

\vspace{1mm}
\noindent\textbf{CIFAR10 Experiments: }
Table~\ref{table:cifar10} shows the experimental results on CIFAR10, including three common neural network models (\textit{i.e.} ConvNet, ResNet18, and DenseNet), as well as three robustness settings (\textit{i.e.} No defense, Gradient Regularization, and MinMax Training).
  The experiment results show that our metric has the same scale with the referenced adversarial testing accuracies, implying our metric's good generality on complex neural network models and different defenses.

\begin{figure}[!b]
  \centering
    \vspace{-3mm}
  \includegraphics[width=3.3in]{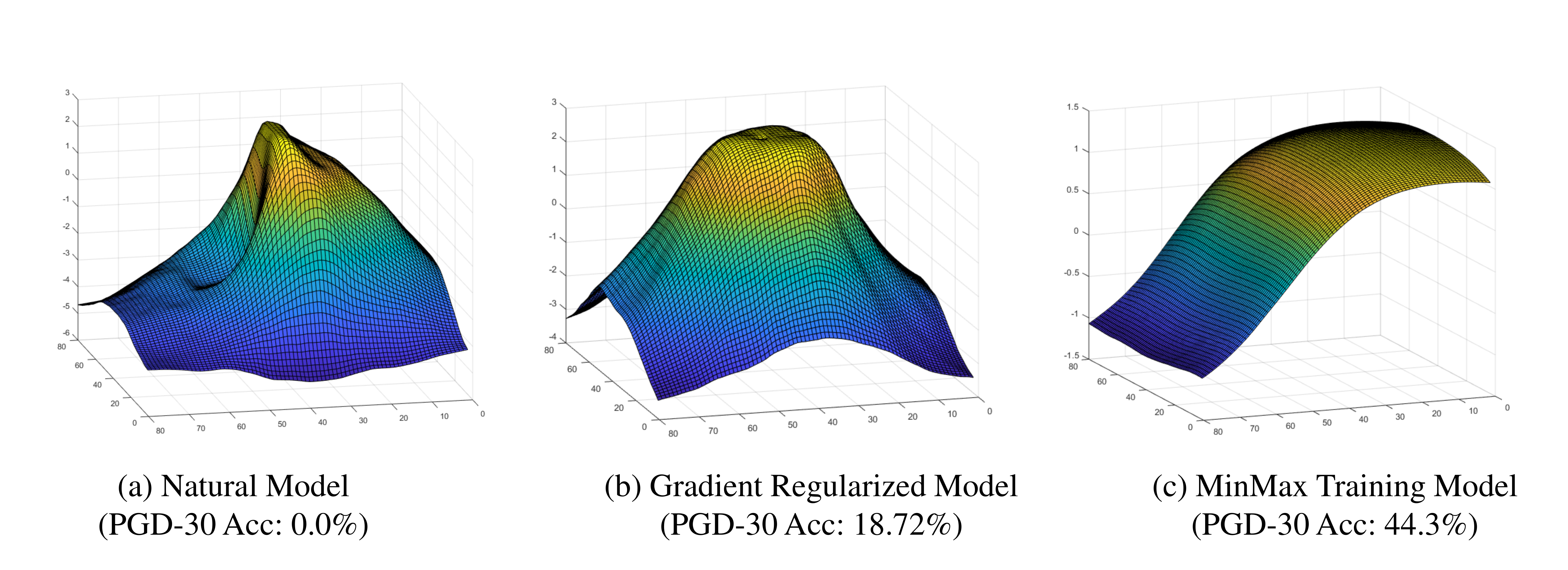}
    \vspace{-2mm}
  \caption{Different models' loss visualizations: model with higher robustness demonstrates more smooth and stable geometry.}
    \vspace{-1mm}
  \label{robust_degrees}
\end{figure}

To better illustrate a neural network model's robustness, we visualized three ResNet18 models with different robustness degrees in Fig.~\ref{robust_degrees}.
    As the robustness degree increases, the models' loss surfaces become more and more smooth.
Our empirical visualization results imply that the smoother decision surface in the input space indicates better adversarial robustness, which coincidentally matches the parameter space generalization hypothesis~\cite{largebatch}.


\vspace{1mm}
\noindent\textbf{ImageNet Experiments: }
  In the experiment on MNIST and CIFAR10, our proposed robustness metric aligns well with adversarial testing accuracies of PGD-30 and CW-30.
  However, when we evaluate the MinMax model on ImageNet, two reference accuracies demonstrate certain inconsistency:

The MinMax model is released as the base model of the state-of-the-art defense on ImageNet by~\cite{imagenet}. To conduct the MinMax training, the reported time needed is about 52 hours on 128 V100 GPUs. Despite that, the reported accuracy showed very good robustness estimation of the model, which can achieve 42.6\%  under 2000-iteration PGD attacks.
    However, when we more thoroughly evaluated the model by CW-30 attack, we found the model's testing accuracy is only 12.5\% under the attack.

We call such a case as  ``\textit{unreliable estimation}'' in PGD-based adversarial testing, whose robustness estimation cannot generalize to all attacks.
  We will discuss this case and several other similar ones in details in Sec.~\ref{sec:failure}, and reveal the current deficiency of adversarial testing based robustness estimation.


\begin{table}[!tb]
  \centering
  \scriptsize
  \renewcommand\arraystretch{1.3}
  \setlength{\tabcolsep}{3.0mm}{
  \caption{Robustness Metric Evaluation on CIFAR10}
  \vspace{-2.5mm}
  \begin{tabular}{ccccc}
    \hline \hline
    Model & Defense & $\psi(x)$ & \begin{tabular}[c]{@{}c@{}}PGD-30\\ Accuracy\end{tabular} & \begin{tabular}[c]{@{}c@{}}C\&W-30\\ Accuracy\end{tabular} \\ \hline \hline
    \multirow{3}{*}{ConvNet} & No Defense & \textbf{58.3} & 0.0\% & 0.0\% \\ \cline{2-5}
    & GradReg & \textbf{86.5} & 16.0\% & 14.8\% \\ \cline{2-5}
    & MinMax & \textbf{182.6} & 39.6\% & 38.7\% \\ \hline
    \multirow{3}{*}{ResNet18} & No Defense & \textbf{67.9} & 0.0\% & 0.0\% \\ \cline{2-5}
    & GradReg & \textbf{77.8} & 18.7\% & 17.5\% \\ \cline{2-5}
    & MinMax & \textbf{162.7} & 44.3\% & 43.1\% \\ \hline
    \multirow{3}{*}{DenseNet} & No Defense & \textbf{59.1} & 0.1\% & 0.0\% \\ \cline{2-5}
    & GradReg & \textbf{77.9} & 18.6\% & 17.2\% \\ \cline{2-5}
    & MinMax & \textbf{142.4} & 39.1\% & 38.8\% \\ \hline
    \hline
  \end{tabular}
*GradReg: \cite{aaai}, MinMax: \cite{minmax}.
  \vspace{-5mm}
\label{table:cifar10}}
\end{table}

\subsection{Our Metric vs. Adversarial Testing Accuracy}
\label{sec:failure}
As mentioned above, the adversarial testing accuracy from different adversarial attacks may demonstrate certain inconsistency, and therefore mislead the robustness estimation.
  In addition to the ImageNet example, we also include another two cases that the adversarial testing accuracies yield unreliable robustness estimation: defensive distillation~\cite{distillation} and gradient inhibition~\cite{wujie}.

To demonstrate the unreliability of these cases, we train three new models on MNIST and CIFAR10 respectively, using natural training, defensive distillation, and gradient inhibition methods.
  For the ImageNet model, we use a public released model \footnote{MinMax model is obtained in following link: https://github.com /facebookresearch/imagenet-adversarial-training.}, which can achieve a state-of-the-art accuracy 45.5\%  against PGD-30 attack (within $\ell_\infty \leq$ 16/255).

The overall experimental results are shown in Table.~\ref{table:advtest}, which shown that though all these defenses can achieve high PGD-30 adversarial testing accuracy, they actually bring very limited robustness improvement:

On MNIST and CIFAR10, the distillation and gradient inhibition defenses provide the models with high adversarial testing accuracy against both FGSM and PGD-30 attacks (even higher than state-of-the-art MinMax methods), which seemly indicates these models are significantly robust.
  However, when measured by our metric, we have the opposite conclusion: these models are merely as robust as no-defense models and incomparable to the robust models trained by MinMax.
  To further verify this conclusion, we test these models with more adversarial settings and the testing accuracy dramatically degrades to almost zero in all the tests.

\begin{table}[!tb]
\centering
\scriptsize
\renewcommand\arraystretch{1.5}
\vspace{1mm}
\caption{Unreliable Cases of Adversarial Testing Accuracy}
\vspace{-2mm}
\begin{tabular}{cccccc}
\hline
\hline
Dataset                  & Defense                & \begin{tabular}[c]{@{}c@{}}FGSM\\ Accuracy\end{tabular} & \begin{tabular}[c]{@{}c@{}}PGD-30\\ Accuracy\end{tabular} & $\psi(x)$ & \begin{tabular}[c]{@{}c@{}}C\&W-30\\ Accuracy\end{tabular} \\ \hline \hline
\multirow{4}{*}{MNIST}   & No Defense              & 23.4\%                                                  & 3.5\%                                                     & 89.8  & 0.5\%                                                    \\ \cline{2-6}
                         & Distillation & \textbf{97.3\%}*                                        & \textbf{97.1\%}*                                           & 70.5  & 0.0\%                                                    \\ \cline{2-6}
                         & GradInhib  & \textbf{98.3\%}*                                         & \textbf{97.8\%}*                                           & 87.0  & 0.0\%                                                    \\ \cline{2-6}
                         & MinMax  & {98.3\%}                                         & {92.3\%}                                           & 958.4  & 90.3\%                                                    \\ \hline
\multirow{4}{*}{CIFAR10} & No Defense              & 7.6\%                                                   & 0.1\%                                                     & 58.3  & 0.0\%                                                       \\ \cline{2-6}
                         & Distillation  & \textbf{72.6\%}*                                        & \textbf{72.3\%}*                                           & 60.5  & 0.0\%                                                     \\ \cline{2-6}
                         & GradInhib  & \textbf{79.8\%}*                                        & \textbf{79.7\%}*                                          & 70.0  & 0.1\%                                                      \\ \cline{2-6}
                         & MinMax  & {55.7\%}                                         & {39.6\%}                                          & 182.6  & 38.7\%                                                    \\ \hline
\multirow{2}{*}{ImageNet}& No Defense           &   15.5\%                                        &   7.3\%                     &  1.7$\times$10$^5$                        & 4.6\%
\\ \cline{2-6}
                         & MinMax   &  \textbf{46.9\%}                                     & \textbf{45.7\%}             &  2.3$\times$10$^5$                        & 12.5\%
                         \\\hline \hline
\end{tabular}
\label{table:advtest}
*Distillation: \cite{distillation}, GradInhib: \cite{wujie},
\vspace{-1mm}
MinMax: \cite{minmax} *Bold accuracies denote the unreliable robustness estimation cases.
\normalsize
\vspace{-2mm}
\end{table}

The tests above further prove our statement: the adversarial testing accuracy based on PGD-30 may yield unreliable robustness estimation, which cannot reflect the model's intrinsic robustness.
   This is because the distillation and gradient inhibition both rely on the input gradient vanishing to achieve robustness enhancement, which is mainly provided by the nonlinear softmax and negative log loss.
   Since C\&W attack doesn't rely on the cross-entropy loss, it can easily crack those two defenses.
  Such a case also applies to the ImageNet model trained with MinMax defenses as shown in the last two rows of Table.~\ref{table:advtest}.

In contrast, our robustness metric can successfully reflect the model true robustness property with different defenses. Under all the above cases, the robustness metric gives reliable robustness estimation, remaining un-affected by defense methods and the unreliable PGD adversarial testing accuracy.

\subsection{Reparemeterization Invariance Evaluation}
The reliability of our proposed metric is also reflected in its invariance from the model parameter scaling.

Previous work~\cite{largebatch} tried to define a metric, named $\epsilon$-sharpness, to evaluate the loss surface's geometry properties. We adopt its original definition and apply it into our input space to evaluate sharpness of input space loss surface, which can empirically reflect the adversarial generalization as aforementioned.

The experiment results are shown in Table.~\ref{table:repar}, where $\epsilon$ denotes the $\epsilon$-sharpness, $\psi_s$ denotes our robustness metric based on softmax layer without normalization, and $\psi_n$ denotes our robustness metric with normalization.
    For the test cases, \textit{Org.} indicates the tests with the original model without reparemeterization, {*100} and {/100} denote the model's logit layer weights and biases are scaled accordingly.
  Please note that, such scaling won't introduce accuracy and robustness change in practice ~\cite{sharpminima}.

The experiments show that, both $\epsilon$-sharpness and un-normalized $\psi_s$ give very distinct robustness estimations influenced by the reparameterization. By contrast, the normalization method successfully alleviates the scaling influence and enables our metric $\psi_n$ to keep a stable estimation under model reparameterization.
Therefore, our metric could thus be used to more precisely capture one model's robustness degree without being affected by model reparameterization.

\begin{table}[!tb]
\centering
\scriptsize
\renewcommand\arraystretch{1.5}
\vspace{1mm}
\caption{Robustness Metrics Comparison under Reparemeterization}
\vspace{-2mm}
\begin{tabular}{ccp{4mm}cp{4.5mm}cp{4.5mm}cccc}
\hline \hline
\multirow{2}{*}{Model}    & \multirow{2}{*}{Metric} & \multicolumn{3}{c}{No Defense Model}                  & \multicolumn{3}{c}{MinMax Model}                        \\ \cline{3-8}
                          &                   & Org.           & *100           & /100           & Org.           & *100            & /100            \\ \hline \hline
\multirow{3}{*}{ConvNet}  & $\epsilon$                 & 22.7           & 109.6          & 0.095          & 0.43           & 3.20            & 0.004           \\ \cline{2-8}
                          & $\psi_s$                & 0.96           & 0.012          & 1677.8         & 39.6           & 5.33            & 377443.3        \\ \cline{2-8}
                          & $\psi_n$                & \textbf{58.3}  & \textbf{59.5}  & \textbf{57.9}  & \textbf{182.5} & \textbf{183.1}  & \textbf{177.36} \\ \hline
\multirow{3}{*}{ResNet18} & $\epsilon$                  & 15.4           & 87.4           & 0.048          & 0.085          & 5.63            & 0.005           \\ \cline{2-8}
                          & $\psi_s$                & 0.963          & 0.0097         & 3178.8         & 17.11          & 0.158           & 128089          \\ \cline{2-8}
                          & $\psi_n$               & \textbf{110.9} & \textbf{110.8} & \textbf{102.5} & \textbf{193.0} & \textbf{192.62} & \textbf{172.5}  \\ \hline
\hline
\end{tabular}
\label{table:repar}
\normalsize
\vspace{-2mm}
\end{table}

\subsection{Efficiency of the Robustness Metric}
\begin{figure}[!b]
    \vspace{-1mm}
  \centering
  \includegraphics[width=3.5in]{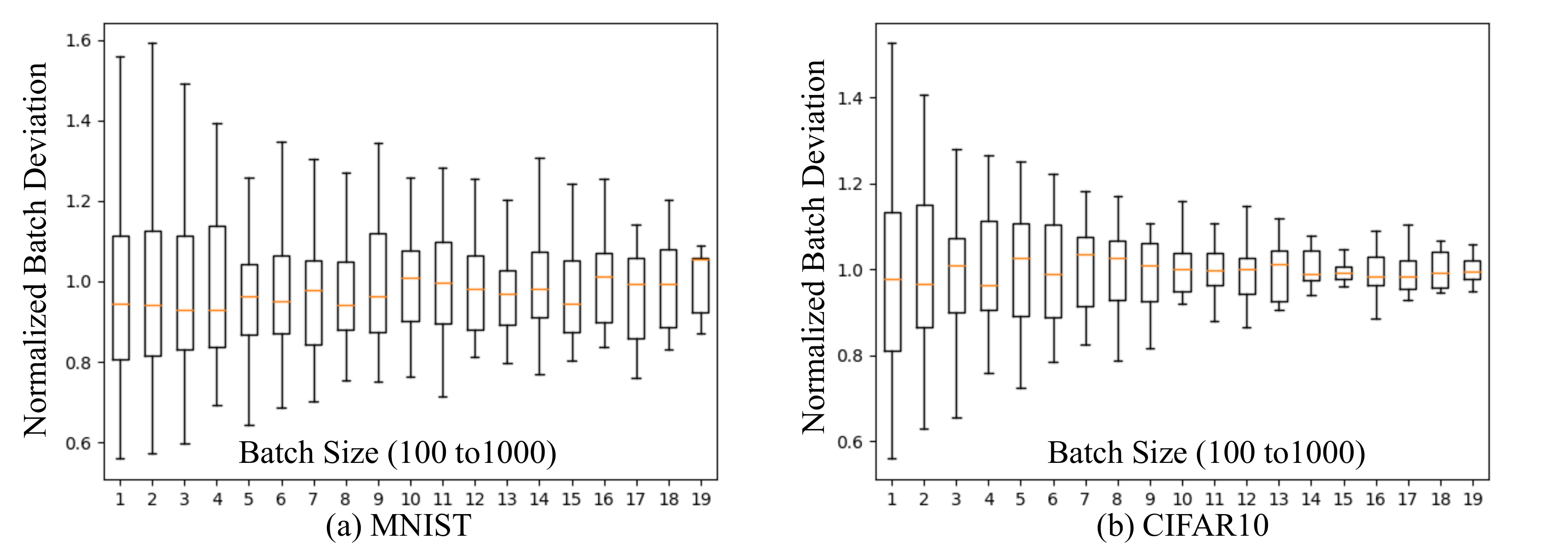}
  \vspace{-5mm}
  \caption{The robustness measurement is increasingly stable with the increasing batch size (100 to 1000).}
  \label{pic:convergence}
  \vspace{-1.5mm}
\end{figure}

Here we show the efficiency of our metric compared to adversarial testing methods. Since we are evaluating the model properties, theoretically it should be invariant to how many input we choose. Here we show that as the test-batch-size increases, the calculated robustness metric gradually converge to a stable robustness estimation which is close to the whole test set average robustness. Fig.~\ref{pic:convergence} shows the relation with the batch size and the robustness deviation between batches with same batch-size. We can see that on both datasets, as the batch size increases, the robustness measurement become more accurate since they have much smaller deviations. With the batch-size equals to 1000 (or less), we could get the model's robustness estimation with less than 10\% deviation on MNIST and 5\% on CIFAR10, which demonstrate higher efficiency than accuracy testing running on the whole test set.
\subsection{Robustness Estimation Grading}

Based on our experiments, we could derive a rough relationship between different robustness evaluation score and the adversarial accuracy. For example, on MNIST dataset within common threat model ($\ell_\infty < 0.3$), we can define the model robustness by three levels: Vulnerable ($acc < 0.3$), Fairly Robust ($0.3 \leq acc < 0.6$) and Robust ($0.6 \leq acc \leq 1.0$). In such a case, the corresponding robust metric range will be ($0,~100$), ($100,~270$), ($270,~\infty$), which could be used to quickly grade a neural network's robustness. The robustness grading for CIFAR and ImageNet cannot be well developed yet due to the limited robustness currently (40\% and 15\%). 


\vspace{-3mm}
\section{Conclusion}
\label{sec:conclusion}

In this work, through visualizing and interpreting neural networks' decision surface in input space, we show that adversarial examples are essentially caused by neural networks' neighborhood under-fitting issue.
	Oppositely, robust models manage to smoothen their neighborhood and relieve such under-fitting effect. Guided by such observation, we propose a model intrinsic robustness evaluation metric based on the model predictions' maximum KL-divergence in a given neighborhood constrain.
	Combined with our new-designed normalization layer, the robustness metric shows multiple advantages than previous methods, including: great generality across dataset/models/attacks/defenses, invariance under reparameterization, and excellent computing efficiency.

\newpage
\bibliographystyle{ijcai19_ref}
\bibliography{ijcai19}

\end{document}